\documentclass[copyright]{eptcs}
\usepackage[utf8]{inputenc}

\usepackage{cmap}
\usepackage[mathcal]{euscript}
\usepackage{listings}
\usepackage{amsfonts}
\usepackage{graphicx}
\usepackage{proof}
\usepackage{url}
\usepackage{centernot}
\usepackage{courier}
\usepackage[ruled,linesnumbered,algo2e]{algorithm2e}

\usepackage[noend]{algpseudocode}
\let\oldReturn\Return
\renewcommand{\Return}{\State\oldReturn}

\usepackage{xcolor}
 \usepackage{float}
\usepackage{mathtools}
\usepackage[framemethod=TikZ]{mdframed}
\usepackage[]{inputenc}
\usepackage[T1]{fontenc}

\usepackage{placeins}
\usepackage{xspace}
\usepackage[font=footnotesize,labelfont=bf]{caption}
\usepackage{subcaption}
\usepackage[framemethod=TikZ]{mdframed}
\usepackage{booktabs}
\usepackage{fancyvrb}
\usepackage{relsize}
\graphicspath{{images/}}
\usepackage{float}
\usepackage{framed}
\usepackage{multirow}
\usepackage{tikz}
\usepackage{pifont}
\usepackage{wrapfig}
\newcommand{\commentD}[1]{}
\newcommand{\fret}{\textsc{FRET}\xspace}
\newcommand{\ros}{\textsc{ROS2}\xspace}
\newcommand{\fretish}{\textsc{FRETish}\xspace}

\newcommand{\commentout}[1]{}

\newcounter{template}

\definecolor{cdarkgreen}{rgb}{0.0,0.4,0.0}
\definecolor{customblue}{rgb}{0.0,0.0,0.7}
\definecolor{cbluegreen}{rgb}{0.0,0.4,0.7}
\definecolor{cpurple}{rgb}{0.5,0.0,0.7}
\definecolor{corange}{rgb}{0.8,0.6,0.2}
\definecolor{cgreen}{rgb}{0,0.6,0}
\colorlet{commentcolour}{green!50!black}
\colorlet{keywordcolour}{magenta!90!black}
\definecolor{MyDarkGreen}{rgb}{0.0,0.4,0.0}
\definecolor{MyBlue}{rgb}{0.0,0.0,0.7}
\definecolor{MyPurple}{rgb}{0.7,0.0,0.7}

\colorlet{punct}{red!60!black}
\definecolor{background}{HTML}{EEEEEE}
\definecolor{delim}{RGB}{20,105,176}
\colorlet{numb}{magenta!60!black}

\newcommand{\bArrow}[1]{\boldsymbol{#1}}

\lstdefinelanguage{smv}{
    literate=
      {->}{{{\color{corange}{$\bArrow{\rightarrow}$}}}}{1}
      {!}{{{\color{corange}{\textbf{!}}}}}{1}
      {\&}{{{\color{corange}{\textbf{\&}}}}}{1}
      {|}{{{\color{corange}{\textbf{|}}}}}{1}
      {=}{{{\color{corange}{\textbf{=}}}}}{1}
      {>}{{{\color{corange}{\textbf{>}}}}}{1}
      {<}{{{\color{corange}{\textbf{<}}}}}{1}
      {:}{{{\color{corange}{\textbf{:}}}}}{1}
      {)}{{{\color{corange}{\textbf{)}}}}}{1}
      {(}{{{\color{corange}{\textbf{(}}}}}{1}
      {?}{{{\color{corange}{\textbf{?}}}}}{1}
      {;}{{{\color{corange}{\textbf{;}}}}}{1}
      {+}{{{\color{corange}{\textbf{+}}}}}{1},
    morekeywords=[1]{LAST,FTP},
    keywordstyle=[1]\color{MyPurple}\bfseries,
    morekeywords=[2]{G, S, H, F, X, Y, U, O, SI},
    keywordstyle=[2]\color{customblue}\bfseries,
    morekeywords=[3]{MODULE,main,VAR,ASSIGN,DEFINE,LTLSPEC,NAME},
    keywordstyle=[3]\bfseries,
    morekeywords=[4]{LAST,MODE,COND,RES},
    keywordstyle=[4]\color{MyPurple}\bfseries,
    morekeywords=[5]{V},
    keywordstyle=[5]\color{corange}\bfseries,
    basicstyle=\scriptsize \ttfamily
}

\definecolor{Mahogany}{rgb}{0.5, 0.0, 0.0}
\definecolor{ForestGreen}{rgb}{0.0, 0.27, 0.13}

\title{Monitoring \ros: from Requirements to Autonomous Robots}
\newcommand{\titlerunning}{Monitoring \ros: from Requirements to Autonomous Robots}
\newcommand{\authorrunning}{Perez, I., Mavridou, A., Pressburger, T., Will, A., Martin, P. J.}
\author{Ivan Perez \qquad\qquad Anastasia Mavridou
\institute{KBR at NASA Ames Research Center}
\and
Tom Pressburger
\institute{NASA Ames Research Center}
\and
Alexander Will \qquad\qquad Patrick J. Martin
\institute{Virginia Commonwealth University}
}

\newcommand{\copyrightholders}{2022. This is a U.S. government work and not under\\ copyright protection in the U.S.; foreign copyright\\ protection may apply.}

\begin{document}

\maketitle

\begin{abstract}

%
Runtime verification (RV) has the potential to enable the safe operation of safety-critical systems that are too complex to formally verify, such as Robot Operating System 2 (ROS2) applications.
%
%
Writing correct monitors can itself be complex, and errors in the monitoring subsystem threaten the mission as a whole.
%
%
This paper provides an overview of a formal approach to generating runtime monitors for autonomous robots from requirements written in a structured natural language.
Our approach integrates the Formal Requirement Elicitation Tool (FRET) with Copilot, a runtime verification framework, through the Ogma integration tool.
FRET is used to specify requirements with unambiguous semantics, which are then automatically translated into temporal logic formul\ae.
Ogma generates monitor specifications from the FRET output, which are compiled into hard-real time C99.
To facilitate integration of the monitors in \ros, we have extended Ogma to generate \ros packages defining monitoring nodes, which run the monitors when new data becomes available, and publish the results of any violations.
The goal of our approach is to treat the generated \ros packages as black boxes and integrate them into larger \ros systems with minimal effort.

\end{abstract}
\section{Introduction}

The Robot Operating System (ROS)~\cite{quigley2009ros} is an ecosystem of software components to support the development of robot applications.
It comprises middleware (e.g., process communication), algorithms (e.g., visual perception, planning), and development tools (e.g., configuration, compilation, visualization, debugging).
Over the last decade, ROS has become one of the most popular frameworks for robotic development, and has been used in manufacturing automation, robot simulations, drones, marine robots, and in assisting astronauts in the International Space Station~\cite{macenski:2022:ros2,Robonaut2,fluckiger:2018:astrobee}.
Recently, a new version, namely \ros, was completely re-engineered to meet the demands of the most stringent applications~\cite{ros:website}. 

When applied in safety-critical domains, \ros systems require high levels of assurance, which may be hard to attain due to the complexity of \ros applications.
That complexity may be inherent to the problem being solved, but it may also stem from the nature of the \ros application architecture itself.
Applications are distributed systems composed of independent \emph{nodes}, which may join or leave dynamically at any point during execution, and communicate via a publish-subscribe mechanism.
The distributed and highly dynamic nature of ROS2 applications and, moreover, the frequent use of AI/ML techniques, increase the difficulty of applying formal verification in this domain~\cite{10.1145/3342355,10.1007/978-3-030-76384-8_4}.

Runtime Verification (RV) is a lightweight verification technique that is suitable for systems that are too complex to formally verify, and can be applied specifically to \ros.
In RV, a system under observation is monitored during runtime, to detect violations of expected properties or requirements.
When a violation takes place, the RV subsystems triggers a notification, which may alert a different part of the system or a human pilot to take action, or log the event for future analysis.

Errors in the RV subsystem can severely impact the mission as a whole.
Depending on the application, the results from the RV monitors may be used to make determinations about the mission, such as aborting the mission completely, or taking corrective actions.
An error in the RV subsystem can therefore affect the complete system and, without additional measures, lead to catastrophic failures.

The correct implementation of runtime monitors for a \ros system can be challenging, due to a number of factors.
The properties that must be monitored may be difficult to specify, especially in low-level programming languages like C++.
Additionally, data passing between the RV subsystem and other parts of a ROS2 system requires substantial boilerplate, which is repetitive and error prone.

This paper proposes a workflow to produce runtime monitoring packages for \ros directly from requirements specified in natural language.
Our proposal relies on three existing tools.
First, FRET is used to specify formal requirements in structured natural language, which automatically translates into temporal logic formul\ae.
Then, the Ogma tool generates a specification of runtime monitors in the stream-based monitoring language Copilot.
Finally, the Copilot compiler produces hard real-time C99 code that implements the monitors specified.

The novel contribution of this work is the extension of Ogma with a prototype backend that generates a \ros monitoring node that wraps the Copilot-generated C monitors.
The code generated passes the input data needed by the monitors when new information arrives, re-evaluates the monitors, and publishes requirement violations.
The output produced by Ogma also includes package descriptions, as well as compilation scripts.
Our goal is to generate self-contained, independent monitoring \ros packages that users can incorporate in their \ros systems without having to modify the automatically generated code.
A prototype implementation has been made available in a branch in Ogma's official repository~\cite{ogma}.
\section{The approach}
\label{sec:approach}

\begin{figure}[t]
\includegraphics[width=\textwidth]{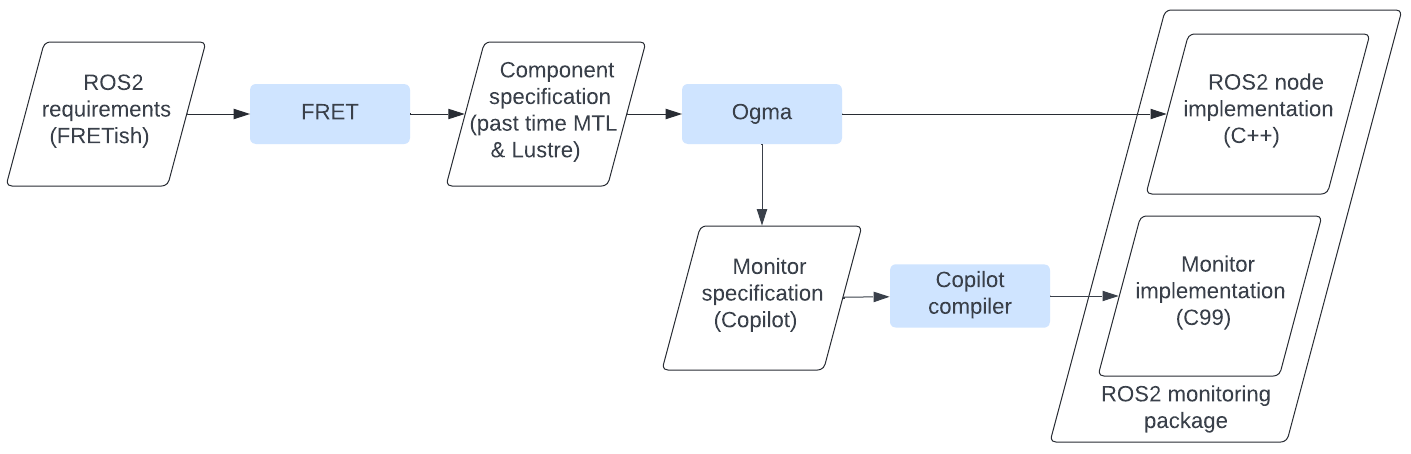}
\caption{\label{fig:workflow} Toolchain to automatically generate monitors for \ros.}
\end{figure}

The process of producing \ros monitoring applications is comprised of three steps~(Fig.~\ref{fig:workflow}).
First, users use the \fret tool to capture requirements in structured natural language as well as generate the formalization of these requirements (\S\ref{sec:fret}).
Then, Ogma generates monitors from FRET's output (\S\ref{sec:monitorgen}).
Third, the monitor information produced by Ogma is sent through a second execution that produces the necessary \ros wrapper node for the monitoring subsystem.
The first two steps of this process predated the current document, and were described in prior work targeting the flight software middleware NASA Core Flight System (cFS)~\cite{perez2021fretcopilot}.
The last step is an extension to Ogma and a novel contribution (\S\ref{sec:rosIntegration}).

To motivate the proposed approach, we consider a scenario relevant for Urban Air Mobility (UAM) vehicles~\cite{cohen2021urban}.
These vehicles feature complex autopilot and autonomy software stacks that make decisions based on multiple internal and external streams of information.
In order to inform mitigation planners and operators of potential hazards, a pervasive monitoring framework would provide a broader view of system health ~\cite{gautham2022pervasive}.
These monitors are specified and deployed at different architectural levels and leverage both internal and external data streams.
For example, one challenge for UAM involves heightened electrical current consumption when a vehicle encounters a high wind zone.
If this high current is sustained too long, it could adversely affect the aircraft's ability to reach its mission destination given the reduced energy reserves.
To evaluate if this hazard is active, we require a monitor that checks whether the current consumption and wind speed exceed desired thresholds for some duration while also checking that this current drops below the threshold after a period of high current draw.

We will assume the UAM vehicle is a quadrotor and the current of each motor is published to a \ros topic.
Additionally, we assume this vehicle has a module to estimate the current wind speed.
We define the following monitoring requirement that captures the sustained high wind, high current scenario:
\begin{quote}
If the current consumption of motor $n$ is greater than some threshold, $cc_t$, for 10 seconds while the wind speed is greater than the threshold $ws_t$, the current consumption shall be less than or equal to $cc_t$ within 10 seconds.   
\end{quote}
In the following sections, this requirement will be mapped through our proposed process and tooling to automatically generate deployable \ros-based monitors.

\subsection{Requirement elicitation and formalization}
\label{sec:fret}

\fret~\cite{fretTool2020} is an open-source tool~\cite{fretgithub} developed at NASA Ames for writing, understanding, formalizing, and analyzing requirements. In practice, requirements are typically written in natural language, which is ambiguous and, consequently, not amenable to formal analysis. Since formal, mathematical notations are unintuitive, requirements in \fret are entered in a restricted natural language named \fretish~\cite{giannakopoulou2021automated}. 

\begin{figure}[t]
\centering
\includegraphics[width=\textwidth]{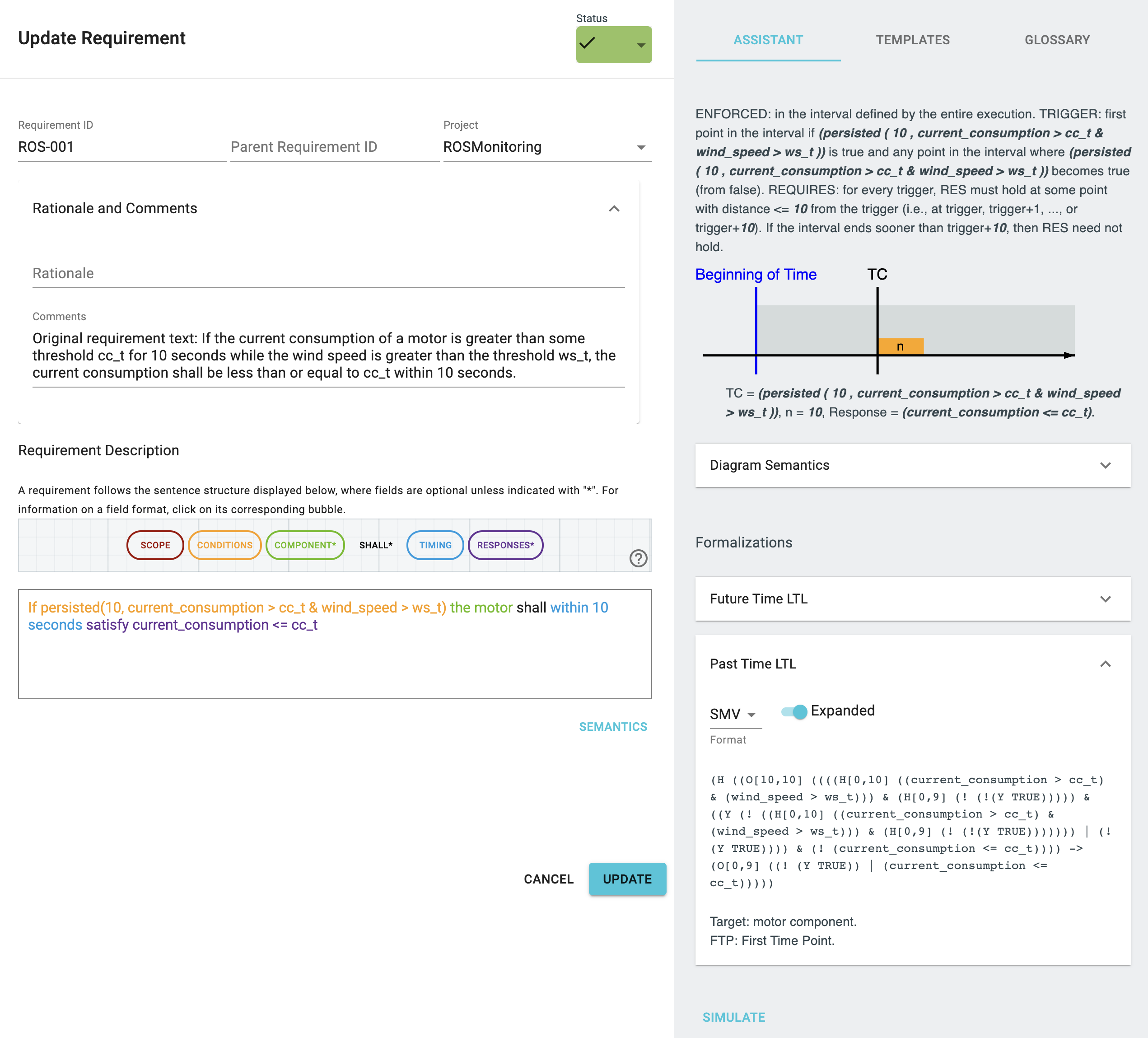}
\caption{\label{fig:ROSFRETrequirement} FRET editor, with the example requirement entered, and the Assistant pane visible.}
\end{figure}

Fig.~\ref{fig:ROSFRETrequirement} illustrates FRET's requirement editor.
The `Comments' field holds the original requirement. 
`Requirement Description' is where the \fretish requirement is composed using up to six distinct fields (the * symbol designates mandatory fields): 1) \textcolor{Mahogany}{\texttt{scope}} specifies the time intervals where the requirement is enforced, 2) \textcolor{orange}{\texttt{condition}} (\textbf{persisted(10, current\_consumption > cc\_t \& wind\_speed > ws\_t)}\footnote{The condition \textbf{persisted(n,p)} becomes true at the time the Boolean expression \textbf{p} has held for the previous \textbf{n} time steps and also holds at the current time step, for a total of \textbf{n+1} time steps, meaning a duration of \textbf{n} time units.}) is a Boolean expression that triggers the \textcolor{violet}{\texttt{response}} to occur at the time the expression's value becomes true, or is true at the beginning of the scope interval, 3) \textcolor{ForestGreen}{\texttt{component*}} (\textbf{ROS\_component}) is the system component that the requirement is levied upon, 4) \texttt{shall*} is used to express that the component's behavior must conform to the requirement, 5) \textcolor{cyan}{\texttt{timing}} (\textbf{within 10 seconds}) specifies when the response shall happen, subject to the constraints defined in \textcolor{Mahogany}{\texttt{scope}} and \textcolor{orange}{\texttt{condition}} and 6) \textcolor{violet}{\texttt{response*}} (\textbf{current\_consumption <= cc\_t}) is the Boolean expression that the component's behavior must satisfy.

Creating a temporal requirement is non-trivial, since such requirements are often riddled with semantic subtleties~\cite{mavridou2020LMCPS}.
To clarify subtle semantic issues, \fret provides a simulator, as well as textual and diagrammatic semantic explanations (Assistant tab in Fig.~\ref{fig:ROSFRETrequirement}).
\fret formalizes \fretish requirements in pure Future-time and pure Past-time Metric Temporal Logic (MTL), and then in Lustre code~\cite{halbwachs1991synchronous}. 
\fret generates the Component specification that is given as input to the Ogma tool.
The Component specification contains the generated specifications in pure Past-time MTL (in SMV format~\cite{bozzano2019nuxmv} and in Lustre code), as well as information about the variables used in the requirements (e.g., data types)~\cite{perez2021fretcopilot}.
 
\subsection{Monitor generation}
\label{sec:monitorgen}

Ogma~\cite{ogma} is a command-line tool that produces monitoring applications based on high-level languages.
Input languages supported by Ogma include FRET component specification format, as well as Lustre node definitions.
The output produced by Ogma contains the implementation for the monitors in Copilot, a stream-based domain-specific language and framework used to specify, simulate and compile monitors~\cite{perez:2020:copilot3}.
Most frequently, Copilot monitors are compiled to hard real-time C99 code, suitable to run on embedded systems.

This two-step process means that users must first use Ogma to generate the Copilot monitors, and then use the Copilot compiler to compile the monitors into C.
This process can easily be automated to generate C directly from \fret component specification files.
In principle, Ogma users do not need to modify the Copilot monitor specifications, although an understanding of Copilot can help debug the monitors individually.

The compilation process through Ogma and Copilot produces a monitor whose status at each point in time can be checked by calling a generated \texttt{step()} function.
Prior to executing that function, the calling function must update the latest values of the inputs needed by the monitors, which are defined as global variables (in our running example, they are \texttt{current\_consumption} and \texttt{windspeed}, both of type \texttt{float}).
After calling \texttt{step()}, the Copilot-generated monitor will report violations by calling a \emph{handler} function, which the user must implement
(in our running example, \texttt{handlerpropROS\_001()}; the name is derived from the requirement ID, \texttt{ROS-001}).
Connecting the generated code to the rest of the application can result in substantial boilerplate.
In the next section we describe how we simplify that process.

\subsection{\ros Integration}
\label{sec:rosIntegration}

\begin{figure}[t]
\begin{center}
\includegraphics[width=0.8\textwidth]{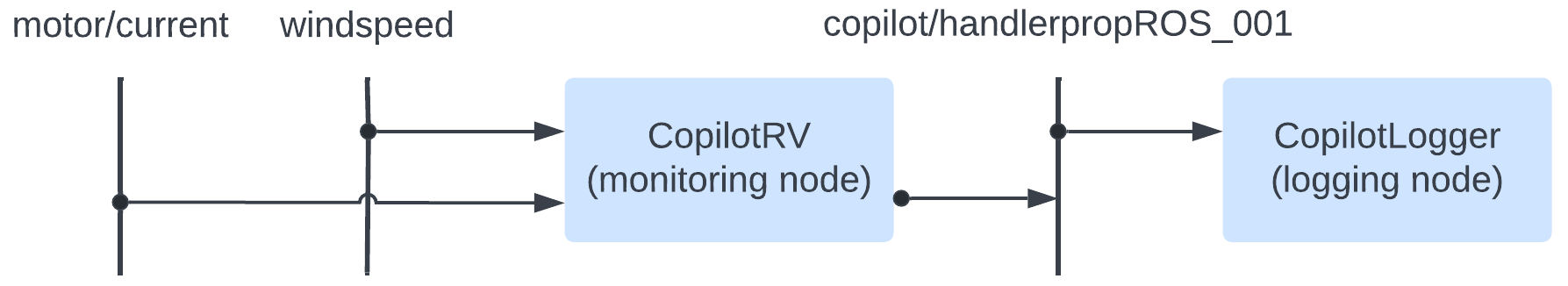}
\caption{\label{fig:archcopilotnode} Architecture of the generated monitoring \ros nodes. Each vertical line represents a \ros topic. Blue boxes represent \ros nodes, arrows pointing from vertical bars to nodes represent subscriptions, and arrows from topics to vertical lines represent messages being published.}
\end{center}
\end{figure}

A novel contribution to Ogma is the generation of the \ros wrapping application that obtains the data needed by the monitors and reports the violations.
Fig. \ref{fig:archcopilotnode} illustrates the automatically generated ROS application based on the use case in Section \ref{sec:approach}.
The monitoring node communicates with other nodes via \emph{topics}, which are strongly typed, named buses.
Any input needed to evaluate the status of the monitors leads to a subscription in the monitoring node to the topic that publishes that data.
When new data arrives on one of the monitoring node's input topics, a handler updates the node's current value of that input, and re-evaluates the monitors.
Conversely, every requirement monitored will result in a separate publisher and a corresponding topic in the monitoring node, which will publish a message every time that requirement is violated.

During the process of generating the \ros monitors, Ogma extracts all variables from the \fret component specification.
Variables represent data sources or inputs for the monitors, so each variable will lead to a subscription to a \ros topic in order to obtain such data.
To produce the necessary subscriptions in the monitoring node, Ogma requires, as additional input, a mapping that relates each variable with its type in \ros and the topic that publishes that data.
For the running example, the inputs would be \texttt{current\_consumption} and \texttt{windspeed}, both of type \texttt{float}, whose new values are obtained from the topics \texttt{motor/current} and \texttt{windspeed}, respectively.
At present, Ogma expects the types and topics associated to variables to be listed in a separate file, which can be shared across a project.

The \ros monitoring application must output any requirement violations.
Ogma associates a topic to each requirement specified in \fret.
In our running example, the only requirement is \texttt{ROS-001}; violations of that requirement are reported as messages published to the topic\linebreak \texttt{copilot/handlerpropROS\_001}.
The name purposefully coincides with the name of the handler declared in the specification of the Copilot monitor automatically generated by Ogma for that requirement.

With the information described, Ogma's \ros backend generates two nodes: a monitoring node, and a logging node~(Fig.~\ref{fig:archcopilotnode}).
The monitoring node works as described before: it subscribes to data sources, re-runs the monitors, and publishes the results.
In our running example, the generated wrapper defines two subscriptions (\texttt{current\_consumption\_subscription\_} and \texttt{windspeed\_subscription\_}), each connected to one topic (respectively, \texttt{motor/current} and \texttt{windspeed}), and one publisher (\texttt{handlerpropROS\_001\_publisher\_}) connected to its own topic\linebreak (\texttt{copilot/handlerpropROS\_001}).
At present, the communication of violations is done via empty messages: their presence indicates a violation, while their absence indicates no violation.
We discuss this further in our future work.

The logging node subscribes to the monitoring node's topics and reports violations to \ros's default logger.
This separate logging node can easily be disabled if not needed, and serves as a suitable starting point to create requirement violation handling nodes.
Ogma also generates the necessary package specification and build scripts.
This application can be embedded into a larger \ros system, and it can be executed by providing the necessary data through the appropriate topics.
\section{Related work}
\label{sec:relatedwork}

ROSRV~\cite{ROSRV} is an RV tool for ROS that inserts a runtime monitoring node, called the RV Master, that intercepts and filters all messages passing through the system.
ROSRV provides a safety property specification language, and automatically generates monitors from specifications.
The tool uses Monitoring-Oriented Programming (MOP~\cite{chen2012semantics}) to specify temporal properties over events.
A difference between ROSRV and our work lies in the expressiveness of the specification language.
ROSRV event handlers, which determine when events must fire, can contain arbitrary programming expressions, and may take arbitrary time to execute.
In contrast, the code Ogma generates for the monitors is hard-realtime and will always finish executing in predictable time.

ROSMonitoring~\cite{ROSMonitoring} is an RV tool for ROS that instruments nodes, intercepting messages at the publisher or subscriber ends.
An oracle verifies properties based on the messages received.
If not all properties are satisfied, the monitor may withhold the message.
ROSMonitoring is formalism-agnostic. It supports Runtime Monitoring Language~\cite{ANCONA2021102610}, along with other formalisms such as Linear Temporal Logic, MTL, and Signal Temporal Logic.
In contrast to ROSMonitoring, we do not provide message filtering capabilities.
ROSMonitoring has been used to synthesize Python monitors from FRET requirements to monitor an autonomous grasping spacecraft motor nozzle for the purposes of debris removal~\cite{farrell:2022:rosmonitoring};
however, the translation from FRET's MTL representation into ROSMonitoring was done manually, whereas our framework generates monitors automatically.

HAROS~\cite{santos:2016:haros} is a framework for quality assurance of ROS systems.
Although the focus of HAROS is static analysis, it is capable of generating runtime monitors, and perform property-based testing.
Like \fretish, the specification language of HAROS introduces notions of contexts or states where the properties must apply, as well as temporal constraints and ranges.
A key difference between our approach and HAROS is that our requirements do not include ROS-specific information, and the process does not require any knowledge of the topology of the ROS graph to generate monitors.
HAROS has also been extended to perform model checking of ROS systems~\cite{carvalho:2020:roselectrum}.
Model checking techniques have also been applied to verify specific ROS systems~\cite{halder:2017:rosuuppal,webster:2015:rosspin}.
HAROS currently only supports ROS1.

DeRoS~\cite{adam:2014:deros} is a domain-specific language and monitoring system specific for ROS.
The language of DeRoS includes explicit notions of topics, while our input language \fretish is platform agnostic, and the connection to ROS is decoupled from the requirements and established via an additional message database provided as input to Ogma.
In addition, DeRoS also allows users to specify how property violations must be handled, while our approach is limited to detecting and reporting violations.

Robonaut2~\cite{Robonaut2} is a humanoid robot running ROS onboard the International Space Station (ISS) performing specialized tasks in collaboration with astronauts.
A fault emerged when Robonaut2 was deployed at the ISS, making the control system freeze and wait for instructions from ground-control at Houston.
Work on the R2U2 runtime monitoring framework~\cite{ISSverification} showed that RV algorithms can efficiently identify faults in real-time.
The application of R2U2 to Robonaut2 was possible, with an adaptation, due to the tool's ability to produce hardware monitors via an FPGA backend that uses size constraints in the monitor synthesis.
In contrast, we produce only software monitors (i.e., in C).

Other runtime monitoring systems, such as Lola~\cite{dangelo:2005:lola}, Java-MOP~\cite{chen:2005:javamop}, detectEr~\cite{aceto:2022:monitoring}, Hydra~\cite{raszyk:2019:multihead,raszyk:2019:hydra}, DejaVu~\cite{havelund:2018:dejavu}, and tracecontract~\cite{barringer:2011:tracecontract}, could in principle also be used for robotics applications.
However, these systems are not made specifically for \ros, and incorporating them in \ros would impose additional development and, potentially, runtime cost.

\section{Summary and Future Work}

We  described a process of generating \ros monitoring packages directly from requirements in structured natural language. Our approach integrates the \fret requirement specification and analysis tool with the Copilot runtime verification framework through Ogma. The generated \ros packages can be integrated into larger \ros systems with minimal effort.

We are currently investigating extensions to \fretish to account for probabilistic requirements, which are useful in autonomous systems.
We are also investigating adding further information to monitor violation messages, such as the values of different variables when violations take place, as well as the time.
Future work includes the following directions.
Currently, the generated nodes re-run all monitors for every input received, which is not always optimal.
It is possible, however, to generate multiple monitoring packages, which will minimize the runtime cost when different requirements are based on largely disjoint sets of variables.
Additionally, we would like to support more evaluation policies (e.g., when all inputs change, on a fixed clock).

\bibliographystyle{eptcs}
\bibliography{paper}
\end{document}